\documentclass[letterpaper, 10 pt, conference]{ieeeconf}  

\IEEEoverridecommandlockouts                              
\usepackage[utf8]{inputenc}
\usepackage[T1]{fontenc}
\usepackage{amsmath,amssymb}
\usepackage{graphicx}
\usepackage{upgreek}
\usepackage{algorithm}
\usepackage{algorithmic}
\usepackage{booktabs}
\usepackage{multirow}
\usepackage{xcolor}
\usepackage{hyperref}
\usepackage[table]{xcolor}
\usepackage{colortbl}
\usepackage{makecell}
\usepackage{array}

\makeatletter
\let\NAT@parse\undefined
\makeatother

\definecolor{myorange}{RGB}{230,120,20}

\hypersetup{
  colorlinks=false,
  pdfborder={0 0 0}
}
\usepackage{microtype}

\newcounter{subfig}[figure]
\renewcommand{\thesubfig}{\alph{subfig}}

\newcommand{\subfiglabel}[2]{%
  \refstepcounter{subfig}%
  \centerline{\scriptsize{(\thesubfig) #1}}%
  \label{#2}%
}

\newcommand{\figsubref}[2]{%
  \hyperref[#2]{\ref*{#1}\ref*{#2}}%
}

\title{\LARGE \bf
VIDP: Variable Impedance Diffusion Policy for Compliant Robot Manipulation from Diverse Demonstrations}

\author{Hisham Khalil$^{1}$, Neil Fernandes$^{2}$, Thomas M. Kwok$^{1}$, Hsiu-Chin Lin$^{3}$, Yue Hu$^{1}$ 
\thanks{This work has been submitted to the IEEE for possible publication. Copyright may be transferred without notice, after which this version may no longer be accessible.}%
\thanks{We acknowledge the support of the Natural Sciences and Engineering Research Council of Canada (NSERC) CREATE ADVENTOR program, funding reference number 584985, and the National Research Council of Canada (NRC), Canada-Japan 3+2 collaborative funding reference number AiP-302-1.}
\thanks{$^{1}$H. Khalil, T. M. Kwok, and Y. Hu are with the Department of Mechanical and Mechatronics Engineering, University of Waterloo, Waterloo, ON, Canada {\tt\small \{hisham.khalil, thomasm.kwok, yue.hu\}@uwaterloo.ca}}
\thanks{$^{2}$N. Fernandes is with the Department of Electrical and Computer Engineering, University of Waterloo, Waterloo, ON, Canada  {\tt\small neil.fernandes@uwaterloo.ca}}
\thanks{$^{3}$H.-C. Lin is with the School of Computer Science and the Department of Electrical and Computer Engineering, McGill University, Montreal, QC, Canada {\tt\small hsiu-chin.lin@mcgill.ca}}
}

\begin{document}
\maketitle
\thispagestyle{empty}

\pagestyle{empty}

\begin{abstract}
Contact-rich manipulation requires precise tracking and mechanical compliance, where variable impedance control can improve robustness in task success, whereas static compliance cannot adapt to varying contact constraints. Variable impedance skills can be learned from demonstrations, avoiding complex modeling, but compliance is a hidden variable in force-agnostic kinematic data. While existing methods infer compliance from trajectory variations, these variations may reflect geometric adaptation and not intentional compliance when subject to changing spatial layouts. Therefore, this letter introduces Variable Impedance Diffusion Policy (VIDP), an imitation learning-based variable impedance control framework leveraging a Task-Parameterized Directionality-Aware Mixture Model (TP-DAMM) to extract physically consistent trajectory distributions from diverse demonstrations. By mapping distributions to stiffness profiles, VIDP jointly predicts pose actions and task compliance without force sensors. Real-world experiments show that VIDP significantly outperforms fixed-impedance baselines in task success rate while reducing interaction forces with respect to high stiffness controllers and tracking errors with respect to low stiffness baselines.
\end{abstract}

\section{Introduction}
\label{sec:introduction}
Collaborative robots performing contact-rich manipulation tasks must balance precise tracking with mechanical compliance to ensure safety \cite{zhang2024srl}, energy efficiency \cite{wu2021learning}, and task success \cite{barreiros2025learning}. However, maintaining static compliance is often insufficient in unstructured environments. For instance, tight-tolerance assembly tasks, such as peg-in-hole insertions, require dynamic stiffness modulation to prevent jamming and accommodate low relative tolerances \cite{abu2020variable}. Similarly, manipulating deformable objects, such as cables or ropes, requires a robot to yield to elastic resistance while applying specific directional forces \cite{lee2015learning}. Therefore, these tasks benefit from variable impedance control, where the level of mechanical compliance varies dynamically over the task horizon.

Learning from demonstrations has emerged as a promising paradigm to transfer variable impedance skills to robots \cite{abu2020variable}, as programming dynamic compliance profiles requires complex analytical modeling of the environment. The fundamental challenge in learning variable impedance control stems from the desired compliance profile, represented as stiffness, being a hidden variable in kinematic demonstrations \cite{kim2021scape}. This limitation becomes a critical issue when attempting to train variable impedance policies that can generalize across changing environments. While providing diverse demonstrations (varying in spatial layouts and object poses) is crucial for task execution generalization, it significantly complicates the estimation of the underlying task compliance. For instance, when a human demonstrates a task across different configurations, the varying trajectories create ambiguity. This creates a challenge for a robot to distinguish between intentional compliance (e.g., yielding to a surface) and spatial deviations (e.g., adapting to a new layout), leading to potentially unpredictable or unsafe physical interactions \cite{romer2025failure}.

Existing imitation learning frameworks attempt to resolve this ambiguity through multi-modal sensing \cite{li2026flow, fang2026force}, which introduces hardware dependencies and synchronization challenges, or by extracting compliance from strictly aligned kinematic data \cite{billard2022learning}, which severely limits spatial generalization. Conversely, standard visuomotor policy learning algorithms like Diffusion Policy \cite{chi2025diffusion} scale well spatially but frame contact-rich manipulation purely as a geometric problem, failing to capture the demonstrator’s underlying physical compliance. Consequently, extracting variable impedance skills from diverse, force-agnostic data remains a critical open challenge.

To address this problem, this letter proposes \textbf{V}ariable \textbf{I}mpedance \textbf{D}iffusion \textbf{P}olicy (VIDP), a framework that predicts compliance profiles from position-only demonstrations by leveraging a Task-Parameterized Directionality-Aware Mixture Model (TP-DAMM). In our approach, TP-DAMM captures the spatial variability of the demonstrations by extracting their task-parameterized covariance profiles. As established in the literature \cite{calinon2010learning, abu2020variable}, these statistical variations directly map to the stiffness parameters of an underlying impedance controller to retrieve task compliance.

The main contributions of this letter are:
\begin{enumerate}
    \item \textbf{A Task-Parameterized Directionality-Aware Mixture Model (TP-DAMM)} that resolves the ambiguity between intentional compliance and geometric adaptation. By extracting physically consistent trajectory distributions from spatially diverse demonstrations, TP-DAMM extracts task variability using solely kinematic data.
    \item \textbf{A Variable Impedance Diffusion Policy (VIDP)} that maps these extracted variabilities into executable target stiffness profiles. This enables a policy to jointly predict precise pose actions and dynamic task compliance from visual and state observations, completely bypassing the need for dedicated force sensors.
\end{enumerate}
We experimentally validate these contributions through quantitative clustering and three real-world contact-rich manipulation tasks. Our results demonstrate that TP-DAMM extracts more physically and directionally consistent motion primitives than standard baselines, while VIDP significantly outperforms fixed-impedance controllers in task success rates, enhancing safety and energy efficiency by minimizing interaction forces, tracking errors, and mechanical work.

\section{Related Work}
\label{sec:related_work}
Classical frameworks for contact-rich manipulation, such as hybrid force/position or operational space impedance control, rely on manual tuning and task-specific modeling, severely limiting scalability across unstructured environments. While reinforcement learning (RL) can adapt stiffness to improve generalization \cite{yang2022variable, karim2025vil}, real-world training remains challenging due to sample inefficiency and exploration safety risks. Consequently, recent work explores learning compliant policies directly from demonstrations. For instance, approaches integrate real-time force feedback with object representations \cite{tsuji2024adaptive}, fuse high-frequency force sensing with vision \cite{he2025foar}, or leverage task-parameterized representations besides force signals to optimize stiffness across various spatial layouts \cite{le2021learning}. While these methods improve task adaptability, they depend on accurate, well-synchronized force/torque measurements, which most commercial collaborative robots lack. Furthermore, these sensors are costly, noisy, and require complex synchronization with low-frequency vision.

Rather than explicitly modeling continuous contact forces, other approaches shift the complexity to the data collection phase. Some constrain teaching by toggling predefined stiffness modes during teleoperation \cite{kamijo2024learning} or utilizing interactive human corrections \cite{xu2025compliant}, demanding extensive manual supervision. Furthermore, while all imitation learning requires structured data collection, existing methods based on diffusion policies for contact-rich manipulation often necessitate highly specialized interactive hardware that limits scalability. For instance, \cite{hou2025adaptive} extracts compliance labels via a dedicated kinesthetic teaching setup. Similarly, \cite{XueH-RSS-25} achieves high precision using a visual-tactile policy that fundamentally relies on fine-grained tactile sensors and complex augmented reality teleoperation rather than standard visual observations.

As an alternative to specialized hardware and manual supervision, stiffness can be estimated directly from demonstrations using force covariance \cite{abu2018force} or inverse position covariance \cite{calinon2010learning}. However, these traditional statistical methods assume consistency within a fixed reference frame, struggling with the unaligned spatial variations necessary to train generalizable policies \cite{knauer2025interactive}. Consequently, a critical gap remains: few approaches integrate compliance directly into visuomotor policy learning without relying on manually tuned gains, multi-modal sensing, or restrictive fixed-frame assumptions. VIDP addresses these limitations by using TP-DAMM to extract physically consistent stiffness profiles from diverse, force-agnostic demonstrations, enabling a visuomotor policy to jointly predict precise pose and compliance.

\section{Background}
\label{sec:background}
\subsection{Classical Variable Impedance Control}
The robot's rigid body dynamics are modeled using an inverse dynamics formulation to compensate for gravity and Coriolis forces \cite{lynch2017modern}. The commanded joint torque $\boldsymbol{\tau}_{cmd}$ follows a hybrid joint impedance control law:
\begin{equation}
\begin{split}
    \boldsymbol{\tau}_{cmd} &= \hat{\mathbf{C}}(\mathbf{q},\dot{\mathbf{q}})\dot{\mathbf{q}} + \hat{\mathbf{g}}(\mathbf{q}) \\
    &\quad + \mathbf{K}_{P}(\mathbf{q})(\mathbf{q}_d - \mathbf{q}) + \mathbf{K}_{D}(\mathbf{q})(\dot{\mathbf{q}}_{d} - \dot{\mathbf{q}}),
\end{split}
\end{equation}
where $\mathbf{q}_{d}$ and $\dot{\mathbf{q}}_{d}$ are the desired joint positions and velocities of a robot with $n$ joints. In variable impedance control, the operational space gains $\mathbf{K}_{d}, \mathbf{D}_{d} \in \mathbb{R}^{6\times 6}$ are dynamically modulated to adapt to task requirements. These gains, along with constant null-space gains $\mathbf{K}_{q}, \mathbf{D}_{q} \in \mathbb{R}^{n\times n}$, are projected through the robot's Jacobian $\mathbf{J}(\mathbf{q})$:
 \begin{equation}
\begin{split}
    \mathbf{K}_{P}(\mathbf{q}) &= \mathbf{J}(\mathbf{q})^{\top} \mathbf{K}_{d} \mathbf{J}(\mathbf{q}) + \mathbf{K}_{q}, \\
    \mathbf{K}_{D}(\mathbf{q}) &= \mathbf{J}(\mathbf{q})^{\top} \mathbf{D}_{d} \mathbf{J}(\mathbf{q}) + \mathbf{D}_{q}.
\end{split}
\end{equation} 
Furthermore, to ensure stable physical interaction, such controllers are frequently maintained at a desired damping ratio $\zeta$ by coupling the operational space damping to the stiffness via $\mathbf{D}_{d} = 2\zeta\sqrt{\mathbf{K}_d}$ \cite{huang2020adaptive}.

\subsection{Task-Parameterized Probabilistic Representations}
\label{sec:tp}
In task-parameterized frameworks, demonstrations are explicitly modeled from the perspective of $P$ candidate reference frames, or task parameters \cite{calinon2013improving}. Let $\boldsymbol{\xi}_t \in \mathbb{R}^D$ denote the global robot state at time $t$. Each frame $j$ is defined by a translation $\mathbf{b}_{t,j}$ and rotation $\mathbf{A}_{t,j}$. The global state is projected simultaneously into all $P$ local frames, generating a dataset of local trajectories $\mathbf{X}_t^{(j)}$:
\begin{equation}
    \mathbf{X}_{t}^{(j)} = \mathbf{A}_{t,j}^{-1}(\boldsymbol{\xi}_{t} - \mathbf{b}_{t,j}).
    \label{eq:tp_projection}
\end{equation}
This isolates the statistical structure of the task relative to specific landmarks, e.g., target objects or robot end-effector. A unified probabilistic model, typically a Task-Parameterized Gaussian Mixture Model (TP-GMM) \cite{calinon2016tutorial}, is learned with parameters $\Theta = \{\pi_m, \{\boldsymbol{\mu}_m^{(j)}, \boldsymbol{\Sigma}_m^{(j)}\}_{j=1}^P\}_{m=1}^M$, sharing mixing coefficients $w_m$ across frames while maintaining separate spatial parameters $(\boldsymbol{\mu}_m^{(j)}, \boldsymbol{\Sigma}_m^{(j)})$ for each frame $j$.

To represent the learned distribution in a new task frame $\{\hat{\mathbf{A}}_{t,j}, \hat{\mathbf{b}}_{t,j}\}_{j=1}^P$, the local densities are projected back into the global frame and fused via a Product of Gaussians (PoG). The fused distribution is given by
\begin{equation}
\hat{\boldsymbol{\Sigma}}_{t,m}
=
\left(
\sum_{j=1}^{P}
\hat{\boldsymbol{\Lambda}}_{t,m}^{(j)}
\right)^{-1},
\quad
\hat{\boldsymbol{\mu}}_{t,m}
=
\hat{\boldsymbol{\Sigma}}_{t,m}
\sum_{j=1}^{P}
\hat{\boldsymbol{\Lambda}}_{t,m}^{(j)}
\hat{\boldsymbol{\mu}}_{t,m}^{(j)},
\end{equation}
where $\hat{\boldsymbol{\mu}}_{t,m}^{(j)}$ and $\hat{\boldsymbol{\Sigma}}_{t,m}^{(j)}$ are the parameters projected into the global frame. This ensures the trajectory satisfies constraints from all task parameters simultaneously.

\begin{figure*}[t]
  \centering
  \includegraphics[width=0.8\linewidth]{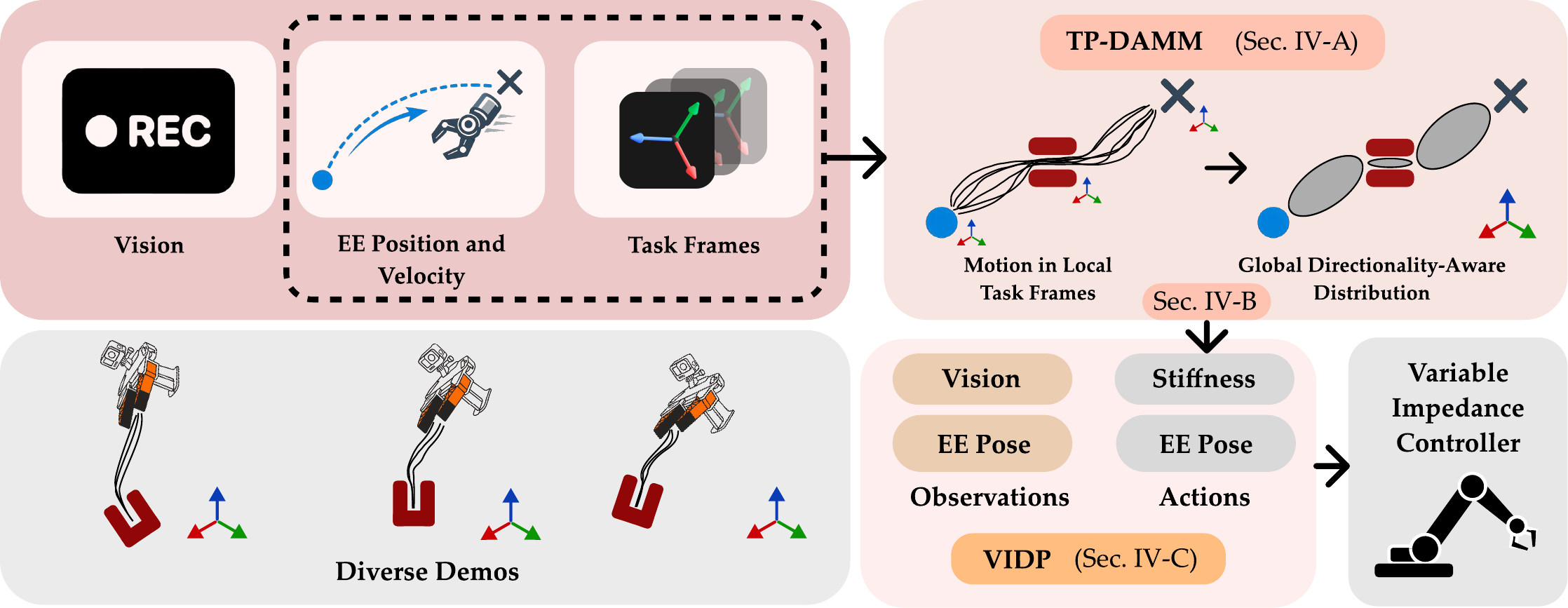}
\caption{\textbf{Overview of VIDP.} (Left) Diverse kinematic demonstrations are collected across varying spatial layouts. (Top Right) TP-DAMM extracts spatially and directionally consistent trajectory distributions. (Bottom Right) Their variances are mapped to stiffness profiles, enabling a Diffusion Policy to jointly predict pose actions and task compliance from real-time visual and state observations for deployment on a variable impedance controller.}
\label{overview}
\vspace{-3mm}
\end{figure*}

While TP-GMMs align spatial distributions across these multiple reference frames, they cannot reliably separate nearby motions with opposing directions. Conversely, Directionality-Aware Mixture Models (DAMM) \cite{sun2024directionality} resolve this directional ambiguity, yet remain confined to a single frame and highly sensitive to global spatial perturbations.
\section{Methodology}
\label{sec:methodology}
VIDP bridges kinematic demonstrations and compliant policy execution through three stages, as shown in Fig. \ref{overview}. First, TP-DAMM extracts spatially and directionally consistent trajectory distributions from diverse demonstrations. Their variance is then mapped to target Cartesian stiffness profiles, which supervise a Diffusion Policy to jointly predict pose actions and compliance from visual and state observations.

\subsection{Trajectory Distribution Estimation via TP-DAMM}
The first stage of our pipeline requires extracting a trajectory distribution that generalizes to diverse environment configurations while preserving the intrinsic dynamical structure of the task. To overcome the spatial limitations of standard DAMM and the directional ambiguity of TP-GMMs discussed in Section \ref{sec:tp}, we propose the TP-DAMM framework. 

\subsubsection{Frame-Specific Directional Augmentation}
Here, $\boldsymbol{\xi}_t \in \mathbb{R}^D$ is instantiated
as the end-effector position, with $\dot{\boldsymbol{\xi}}_t$ its velocity. We model the task using $P$ candidate task frames, where at time $t$, each frame $j$ is defined by a translation $\mathbf{b}_{t,j}$ and a rotation $\mathbf{A}_{t,j}$.

To capture the directional features missing in standard TP-GMMs, TP-DAMM augments the state with a directional deviation term derived from Riemannian statistics on the unit sphere $\mathcal{S}^{d-1}$. While positions transform via the full affine definition in Eqn. \eqref{eq:tp_projection}, the directionality term must be transformed purely via rotation to preserve the unit-norm geometry. For frame $j$, the local position $\boldsymbol{x}_{t}^{(j)}$ and local unit velocity direction $\boldsymbol{u}_{t}^{(j)}$ are computed as:
\begin{equation}
    \boldsymbol{x}_{t}^{(j)} = \mathbf{A}_{t,j}^{-1}(\boldsymbol{\xi}_{t} - \mathbf{b}_{t,j}), \quad \boldsymbol{u}_{t}^{(j)} = \frac{\mathbf{A}_{t,j}^\top \dot{\boldsymbol{\xi}}_t}{\left\lVert \mathbf{A}_{t,j}^\top \dot{\boldsymbol{\xi}}_t \right\rVert}.
\label{eqn:transform}
\end{equation}

Per mixture component $m$, the local directional mean $\boldsymbol{\mu}_{dir,m}^{(j)} \in \mathcal{S}^{d-1}$ is computed as the Fréchet mean of the unit velocity directions assigned to that component. The local directional variance $d_{t,m}^{(j)}$ is then computed via the Riemannian logarithmic map:
\begin{equation}
    d_{t,m}^{(j)} = \left\lVert \text{log}_{\boldsymbol{\mu}_{dir,m}^{(j)}}(\boldsymbol{u}_{t}^{(j)}) \right\rVert.
    \label{eq:dir_var}
\end{equation}

Intuitively, $d_{t,m}^{(j)}$ measures the angular deviation of a
demonstrated velocity direction from its cluster's characteristic
direction, allowing TP-DAMM to distinguish co-located motions that move in opposite directions. The resulting augmented state vector for frame $j$ and component $m$ is given as:
\begin{equation}
    \hat{\boldsymbol{x}}_{t,m}^{(j)} = \left[ \boldsymbol{x}_{t}^{(j)\top},  d_{t,m}^{(j)}\right]^\top \in \mathbb{R}^{d+1}.
        \label{eq:aug_state}
\end{equation}
Consequently, the subsequent probabilistic clustering operates directly on this augmented state, combining the spatial positions and velocity directionality into a unified representation.
\subsubsection{Probabilistic Formulation} 
We model the demonstrations as a mixture of $M$ components, where each component $m$ is represented by a set of Gaussians $\{\mathcal{N}(\boldsymbol{\mu}_m^{(j)}, \boldsymbol{\Sigma}_m^{(j)})\}_{j=1}^P$. Referring to DAMM structure, the covariance matrix $\boldsymbol{\Sigma}_m^{(j)}$ is block-diagonal to enforce independence between the spatial distribution and the directional variance $(\sigma^2)_{dir, m}^{(j)}$:
\begin{equation}
    \boldsymbol{\Sigma}_m^{(j)} = \begin{bmatrix} \boldsymbol{\Sigma}_{pos, m}^{(j)} & \mathbf{0} \\ \mathbf{0} & (\sigma^2)_{dir, m}^{(j)} \end{bmatrix}, \quad \boldsymbol{\mu}_m^{(j)} = \begin{bmatrix} \boldsymbol{\mu}_{pos, m}^{(j)} \\ 0 \end{bmatrix}.
\label{eq:block_diag}
\end{equation}
Let $z_t \in \{1, \dots, M\}$ denote the latent assignment variable indicating that the observation at time $t$ belongs to mixture component $m$. The probability of observing a data-point $\boldsymbol{\xi}_t$ given this assignment is determined by the Product of Gaussians (PoG) across all frames:
\begin{equation}
    \mathcal{P}(\boldsymbol{\xi}_t | z_t=m) \propto \pi_m \prod_{j=1}^P \mathcal{N}\left(\hat{\boldsymbol{x}}_{t,m}^{(j)} \middle| \boldsymbol{\mu}_m^{(j)}, \boldsymbol{\Sigma}_m^{(j)} \right).
\label{eq:pog_likelihood}
\end{equation}

\subsubsection{Parallel Inference with Split/Merge Proposals} 
We use a parallel split-merge Gibbs sampler for Bayesian inference. Unlike standard DAMM's single global representation, TP-DAMM independently samples local position covariances $\boldsymbol{\Sigma}_{pos,m}^{(j)}$ and directional variances $(\sigma^2)_{dir,m}^{(j)}$ per component $m$ and frame $j \in \{1,\dots,P\}$ via Normal-Inverse-Wishart (NIW) and Inverse-Gamma (IG) priors, respectively. To account for multi-frame likelihoods, the inference process alternates between parameter sampling, Gibbs assignment, and split/merge proposals, as shown in Algorithm \ref{alg:tp_damm}.

Furthermore, while DAMM evaluates split proposals using single-frame position data, TP-DAMM utilizes a task-parameterized augmented space. This space concatenates local positions and directional scalars from all $P$ frames, yielding $\mathbf{X}_m \in \mathbb{R}^{N_m \times D_{aug}}$ ($D_{aug}=P(N+1)$). To prevent degenerate covariance estimation when $N_m<D_{aug}$, we apply Principal Component Analysis. By defining the effective dimensionality $d_{\mathrm{eff}}=\min(D_{aug},\max(2,N_m-1))$ and computing the Singular Value Decomposition of the centered data $\mathbf{\bar{X}}_m = \mathbf{U}\mathbf{S}\mathbf{V}^\top$, we project the observations onto the first $d_{\mathrm{eff}}$ principal components $\mathbf{V}_{\mathrm{eff}}$. This allows safe estimation of split parameters in the reduced space as $\mathbf{Z}_m = \mathbf{\bar{X}}_m \mathbf{V}_{\mathrm{eff}}$.

\begin{algorithm}[t]
\caption{TP-DAMM Inference}
\label{alg:tp_damm}
\begin{algorithmic}[1]
\small
\REQUIRE Trajectories $\{\boldsymbol{\xi}_t, \dot{\boldsymbol{\xi}}_t\}_{t=1}^T$, Frames $\{\mathbf{b}_{t,j}, \mathbf{A}_{t,j}\}_{j=1}^P$
\STATE Compute local $\boldsymbol{x}_{t}^{(j)}, \boldsymbol{u}_{t}^{(j)}$ (Eqn. \eqref{eqn:transform}) and initialize $z_t$
\FOR{iteration $= 1$ to $N_{iter}$}
    \STATE \texttt{// 1. Parameter Sampling}
    \FOR{each $m, j$}
        \STATE Sample $\{\boldsymbol{\mu}_{pos}, \boldsymbol{\Sigma}_{pos}\}_m^{(j)} \sim \text{NIW}$, $(\sigma^2)_{dir, m}^{(j)} \sim \text{IG}$
        \STATE $\boldsymbol{\mu}_{dir, m}^{(j)} \leftarrow \text{FréchetMean}(\{\boldsymbol{u}_t^{(j)} : z_t=m\})$
        \STATE Construct augmented $\boldsymbol{\mu}_m^{(j)}, \boldsymbol{\Sigma}_m^{(j)}$ (Eqn. \eqref{eq:block_diag})
    \ENDFOR

    \STATE \texttt{// 2. Gibbs Assignment}
    \FOR{each $t$}
        \STATE Construct augmented state $\hat{\boldsymbol{x}}_{t,m}^{(j)}$ (Eqns. \eqref{eq:dir_var} and \eqref{eq:aug_state})
        \STATE Sample $z_t$ using PoG likelihood (Eqn. \ref{eq:pog_likelihood})
    \ENDFOR

    \STATE \texttt{// 3. Split/Merge Proposals}
    \STATE Build augmented space $\mathbf{X}_m$, center data to $\mathbf{\bar{X}}_m$
    \STATE SVD projection to $d_{\mathrm{eff}}$: $\mathbf{Z}_m = \mathbf{\bar{X}}_m \mathbf{V}_{\mathrm{eff}}$ ($\mathbf{\bar{X}}_m = \mathbf{U}\mathbf{S}\mathbf{V}^\top$)
    \STATE Propose $\Theta_{new}$ (split/merge); Compute acceptance ratio $a$
    \IF{$\log(\text{Uniform}(0,1)) < \log a$} 
        \STATE Accept $\Theta_{new}$, update $z_t$ 
    \ENDIF
\ENDFOR
\ENSURE $\Theta = \{w_m, \{\boldsymbol{\mu}_m^{(j)}, \boldsymbol{\Sigma}_m^{(j)}\}_{j=1}^P\}_{m=1}^M$
\end{algorithmic}
\end{algorithm}

\subsection{Mapping Distributions to Target Stiffness}
\label{stiffness_estimation}
To regulate the robot's compliance, we adopt the learning-based control strategy proposed in \cite{calinon2010learning}, which exploits the task redundancy observed across multiple demonstrations. As motivated in Section \ref{sec:introduction}, the core principle is that the variability of the demonstrations encapsulates the task constraints, in which parts of the motion with high variability indicate low precision requirements (lower stiffness), while parts with low variability require precise tracking (higher stiffness).

Given the fused spatial covariance matrix $\hat{\boldsymbol{\Sigma}}_k$ derived from the TP-DAMM for the $m$-th component, we estimate the translational stiffness matrix $\mathbf{K}_{trans} \in \mathbb{R}^{3\times3}$ by analyzing the inverse covariance matrix \cite{calinon2010learning}. We first perform an eigen decomposition of the inverse covariance:
\begin{equation}
    (\hat{\boldsymbol{\Sigma}}_m)^{-1} = \boldsymbol{V}_m \boldsymbol{\Lambda}_m \boldsymbol{V}_m^{\top},
\label{eq:eigen_decomp}
\end{equation}
where $\boldsymbol{V}_m$ is the matrix of eigenvectors acting as the principal axes of stiffness, and $\boldsymbol{\Lambda}_m = \text{diag}(\lambda_{m,1}, \dots, \lambda_{m,d})$ are the eigenvalues representing the magnitude along those axes.

To map these values to executable stiffness gains, we define a diagonal matrix of stiffness eigenvalues $\boldsymbol{D}_m$. Following the strategy in \cite{calinon2010learning}, we rescale the statistical eigenvalues $\lambda_{m,i}$ to lie within the robot's admissible stiffness range $[\kappa_{\min}, \kappa_{\max}]$:
\begin{equation}
    \boldsymbol{D}_m^{(i,i)} = \kappa_{\min} + (\kappa_{\max} - \kappa_{\min}) \frac{\lambda_{m,i} - \lambda_{\min}}{\lambda_{\max} - \lambda_{\min}},
\label{eq:stiffness_scaling}
\end{equation}
where $\lambda_{\min}$ and $\lambda_{\max}$ are the minimum and maximum eigenvalues observed across the entire task, i.e., across all $M$ components. This ensures that the components with the highest precision requirements are assigned the maximum available stiffness $\kappa_{\max}$, while the most variable components are assigned $\kappa_{\min}$. The stiffness matrix $\mathbf{K}_{trans}^m$ for component $m$ is then reconstructed as $\mathbf{K}_{trans}^m = \boldsymbol{V}_m \boldsymbol{D}_m \boldsymbol{V}_m^{\top}$. To prevent discontinuous control inputs during cluster transitions, which can cause instability in the variable impedance controller \cite{kronander2016stability}, we apply a temporal moving average filter to smoothly interpolate the final reference variable stiffness $\mathbf{K}_{trans,t} = \mathrm{diag}(K_{x}, K_{y}, K_{z})$ along the trajectory.

\subsection{Policy Learning and Real-Time Execution}
\subsubsection{Policy Architecture and Dataset Formulation}
 We formulate VIDP as a Conditional Denoising Diffusion Probabilistic Model (DDPM) \cite{chi2025diffusion}. Because kinematic demonstrations lack force data, we first infer variable stiffness profiles $\mathbf{K}_{trans,t}$ by processing the demonstration trajectory data with TP-DAMM (Section \ref{stiffness_estimation}). This yields a training dataset $\mathcal{D} = \{(\mathbf{O}_t, \mathbf{A}_t)\}$, where $\mathbf{O}_t$ contains visual and proprioceptive observations, and $\mathbf{A}_t$ includes target poses and stiffness parameters. The policy $\pi_\theta$ takes an observation of horizon $T_{obs}$ as input and predicts an action chunk $\mathbf{A}_{pred}$ of horizon $T_{pred}$. Note that VIDP predicts only translational stiffness, sufficient for our translation-dominated evaluation tasks. 

\textbf{Observation Space.} The policy condition $\mathbf{O}_t$ consists of visual and proprioceptive observations. For visual observations, we utilize raw RGB images from the wrist-mounted camera. We then employ a CLIP-pretrained Vision Transformer (ViT) backbone to extract semantic features. To adapt to our specific domain, we fine-tune the encoder end-to-end while training our policy and use attention pooling to aggregate the patch embeddings into a compact feature vector. For proprioception, the robot's end-effector pose history is represented by its 3D position and a continuous 6D rotation representation used to avoid representation singularities during training \cite{zhou2019continuity}. The history is flattened and projected to a feature embedding using a Multi-Layer Perceptron.

\textbf{Action Space.} We augment the kinematic action space to explicitly control compliance. The predicted action chunk $\mathbf{A}_{init} \in \mathbb{R}^{T_{pred} \times 15}$ consists of a sequence of steps $\mathbf{a}_t = [ \mathbf{p}_t^\top, \mathbf{r}_t^\top, \mathbf{c}_t^\top ]^\top$ for $t=1 \dots T_{pred}$, where $\mathbf{p}_t \in \mathbb{R}^3$ is the target position and $\mathbf{r}_t \in \mathbb{R}^6$ is the continuous rotation representation. For the stiffness parameters, simply predicting a full stiffness matrix from the policy does not guarantee symmetric positive definite properties. To address this, we adopt the Cholesky decomposition representation \cite{abu2018force}. We predict the vector $\mathbf{c}_t \in \mathbb{R}^6$, which corresponds to the flattened non-zero elements of the upper triangular Cholesky factor $\mathbf{U}_t$. The full stiffness matrix is reconstructed during execution as $\mathbf{K}_{trans,t} = \mathbf{U}_t^\top \mathbf{U}_t$.

\textbf{Training Objective.} We use a 1D Temporal U-Net as the noise prediction network $\boldsymbol{\epsilon}_\theta$. The network is trained using the AdamW optimizer with a cosine learning rate scheduler. During training, the model minimizes the Mean Squared Error (MSE) between the sampled noise $\boldsymbol{\epsilon}$ and the predicted noise:
\begin{equation}
    \mathcal{L} = \mathbb{E}_{k, \mathbf{A}_0, \boldsymbol{\epsilon}} \left[ \| \boldsymbol{\epsilon} - \boldsymbol{\epsilon}_\theta(\mathbf{A}_k, k, \mathbf{O}_t) \|^2 \right].
\end{equation}

\begin{table*}[t]
\centering
\caption{Quantitative evaluation of clustering performance on three contact-rich task 3D motion datasets. $\uparrow$ indicates higher is better and $\downarrow$ indicates lower is better. Best results are bolded.}
\vspace{-2mm}
\label{tab:compact_all_task_clustering}
\large
\setlength{\tabcolsep}{3.5pt}
\renewcommand{\arraystretch}{1.12}
\resizebox{\textwidth}{!}{
\begin{tabular}{lcccccccccccc}
\toprule
\multirow{2}{*}{\textbf{Method}}
&
\multicolumn{4}{c}{\textbf{Peg-in-Hole Insertion}}
&
\multicolumn{4}{c}{\textbf{Pulley Assembly}}
&
\multicolumn{4}{c}{\textbf{Cable Routing}}
\\
\cmidrule(lr){2-5}
\cmidrule(lr){6-9}
\cmidrule(lr){10-13}
&
\textbf{Loc. Dir. Var.} $\downarrow$
& \textbf{Glob. Dir. Var.} $\downarrow$
& \textbf{Cos.} $\uparrow$
& \textbf{Cover.} $\uparrow$
&
\textbf{Loc. Dir. Var.} $\downarrow$
& \textbf{Glob. Dir. Var.} $\downarrow$
& \textbf{Cos.} $\uparrow$
& \textbf{Cover.} $\uparrow$
&
\textbf{Loc. Dir. Var.} $\downarrow$
& \textbf{Glob. Dir. Var.} $\downarrow$
& \textbf{Cos.} $\uparrow$
& \textbf{Cover.} $\uparrow$
\\
\midrule

GMM
& 0.980 & 0.976 & 0.607 & 0.523
& 1.199 & 1.199 & 0.562 & 0.464
& 1.924 & 1.925 & 0.289 & 0.822
\\

DAMM
& 0.971 & 0.958 & 0.617 & 0.917
& 0.676 & 0.663 & 0.722 & 0.727
& 0.999 & 0.999 & 0.598 & 0.975
\\

TP-GMM
& 0.952 & 0.947 & 0.623 & 0.781
& 0.726 & 0.722 & 0.732 & \textbf{1.000}
& 1.028 & 1.028 & 0.601 & 0.999
\\

\midrule

\rowcolor{gray!12}
TP-DAMM (Ours)
& \textbf{0.943} & \textbf{0.935} & \textbf{0.628} & \textbf{0.933}
& \textbf{0.581} & \textbf{0.574} & \textbf{0.781} & \textbf{1.000}
& \textbf{0.845} & \textbf{0.846} & \textbf{0.662} & \textbf{1.000}
\\
\bottomrule 
\noalign{\vspace{1.5mm}}
\multicolumn{13}{@{}l}{\large Loc. Dir. Var. and Glob. Dir. Var. denote Local and Global Directional Variance, respectively; Cos. denotes Cosine Similarity; Cover. denotes Cluster Coverage.}
\end{tabular}
}
\vspace{-4mm}
\end{table*}

\subsubsection{Real-Time Policy Optimization}
To accelerate real-time inference, we use the Denoising Diffusion Implicit Model (DDIM) scheduler \cite{chi2025diffusion}. However, synchronous Diffusion Policy execution still introduces latency-induced pauses between action chunks, causing dangerous force spikes and jerky transitions in contact-rich manipulation. We therefore employ Real-Time Chunking (RTC) \cite{black2025realtime}, which asynchronously decouples inference from execution to maintain reactive control. To adapt our trained diffusion model to RTC's flow-based guidance, we reinterpret it as an Optimal Transport conditional flow during inference \cite{tong2024improving}. This enables gradient-based trajectory editing to enforce temporal consistency between consecutive action chunks:
\begin{equation}
\hat{\mathbf{v}} = \mathbf{v}_\theta + \gamma(\tau) \nabla_{\mathbf{z}_\tau} \| \mathbf{W} \odot (\mathbf{A}_{prev} - \hat{\mathbf{A}}_0(\mathbf{z}_\tau)) \|^2,
\end{equation}
where $\tau \in [0,1]$ denotes the continuous flow time during inference, while $\gamma(\tau)$ and $\mathbf{W}$ are time-dependent guidance weights and soft-masking vectors, respectively. This mechanism ensures strict consistency for immediate actions while allowing the policy to relax constraints for future steps, enabling smooth reactions to new observations $\mathbf{O}_t$.

\section{Experimental Evaluation}
\label{sec:exp_eval}
We aim to answer the following research questions:
\begin{itemize}
    \item \textbf{RQ1:} Does TP-DAMM extract more directionally consistent and representative task trajectory distributions from spatially diverse demonstrations compared to standard clustering methods?
    \item \textbf{RQ2:} Does VIDP outperform fixed-impedance baselines in success rate across tasks characterized by varying contact constraints and diverse spatial layouts?
    \item \textbf{RQ3:} How does the inferred variable impedance profile optimize the physical interaction metrics, specifically interaction forces, tracking errors, and energy efficiency, during critical execution phases?
\end{itemize}

\subsection{Experimental Setup}
\subsubsection{Hardware}
We collect demonstrations using the Universal Manipulation Interface (UMI) \cite{chi2024universal}, with egocentric GoPro Hero 11 vision and Vicon pose tracking. Deployment uses a Franka Robotics Panda manipulator in a 65 $\times$ 75 cm workspace with custom 3D-printed fingers (see Fig. \ref{fig:stiffness}). Models are trained on an NVIDIA RTX 4090 workstation and run in real time on a laptop with an AMD Ryzen 7 6800H CPU and NVIDIA RTX 3070 Ti GPU.
\subsubsection{Evaluation Tasks and Datasets}
\label{tasks}
We evaluate VIDP on three contact-rich manipulation tasks (see Fig. \ref{fig:stiffness}). To perform thorough task success evaluation, we divide executions into semantic phases defined by transitions in physical interaction states or key spatial waypoints. The tasks include: \textit{Peg-in-Hole Insertion} (333 demonstrations) for rigid contact (Reach, Align, and Insert phases); Pulley Assembly (319 demonstrations) for hybrid interaction (Stretch, Route, and Secure phases); and Cable Routing (307 demonstrations) for deformable manipulation (three sequential Hook phases). Dataset sizes vary due to the post-processing removal of human execution errors or Vicon data noise. To capture spatial variability in TP-DAMM with task frames, \textit{Peg-in-Hole Insertion} uses the start pose, representing the end-effector holding the peg, and the hole location. \textit{Pulley Assembly} uses the start pose, end pose, and pulley location, while \textit{Cable Routing} uses the three routing hooks' locations.

\subsubsection{Baselines}
We compare VIDP with two baselines sharing the same Diffusion Policy \cite{chi2025diffusion} but using fixed-impedance control: Stiff-DP and Compliant-DP, with Cartesian stiffness set uniformly across all axes to 1200 N/m for rigid interactions and 400 N/m for softer contact, respectively. For all deployments, our variable impedance controller is maintained at near-critical damping ($\zeta = 0.707$), while the rotational and null-space stiffness parameters are held fixed at conservative nominal values.

\subsection{TP-DAMM Evaluation}
To answer \textbf{RQ1}, we compare the clustering performance of TP-DAMM against baseline methods: standard GMM, DAMM \cite{sun2024directionality}, and TP-GMM \cite{calinon2016tutorial}.

\subsubsection{Qualitative Evaluation on 2D PC-GMM Dataset}
\label{sec:pc_gmm}
\begin{figure}[t]
  \centering
  \includegraphics[width=1.0\linewidth]{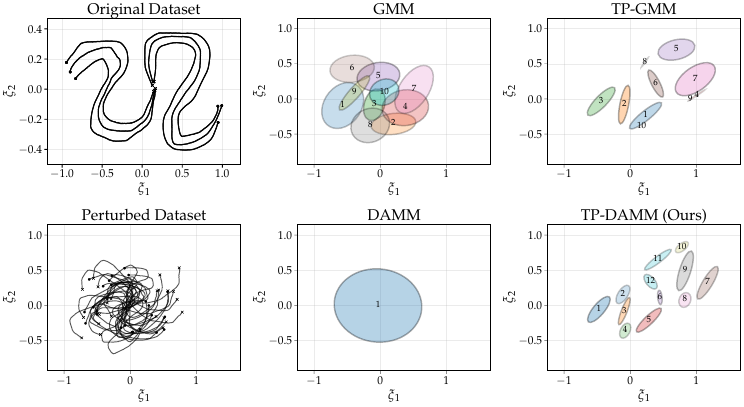}
  \vspace{-6mm}
\caption{Visual comparison of clustering methods on a spatially perturbed 2D PC-GMM dataset. Standard GMM merges opposing motions, while DAMM fails to adapt to global spatial shifts. TP-GMM aligns frames but cannot separate overlapping opposing directions. Conversely, TP-DAMM successfully preserves both the task-frame spatial structure and local motion directionality, closely estimating the original unperturbed task distribution.}
\label{tpdamm_qualitative}
\end{figure}

Fig. \ref{tpdamm_qualitative} illustrates the challenge of clustering trajectories with shifting reference frames and opposing motions using the 2D PC-GMM benchmark dataset \cite{figueroa2018physically}. Since this dataset originally uses a single global frame, we inject candidate reference frames and apply spatial perturbations to observe how effectively the models adapt to varying spatial layouts. The baselines exhibit several limitations. GMM merges opposing motions, DAMM is sensitive to global spatial shifts, and TP-GMM cannot distinguish locally overlapping but opposing motions. Conversely, TP-DAMM successfully preserves both the spatial structure and local directionality.

\begin{figure*}[t]
  \centering
  \begin{minipage}[t]{0.322\linewidth}
    \centering
    \includegraphics[width=\linewidth]{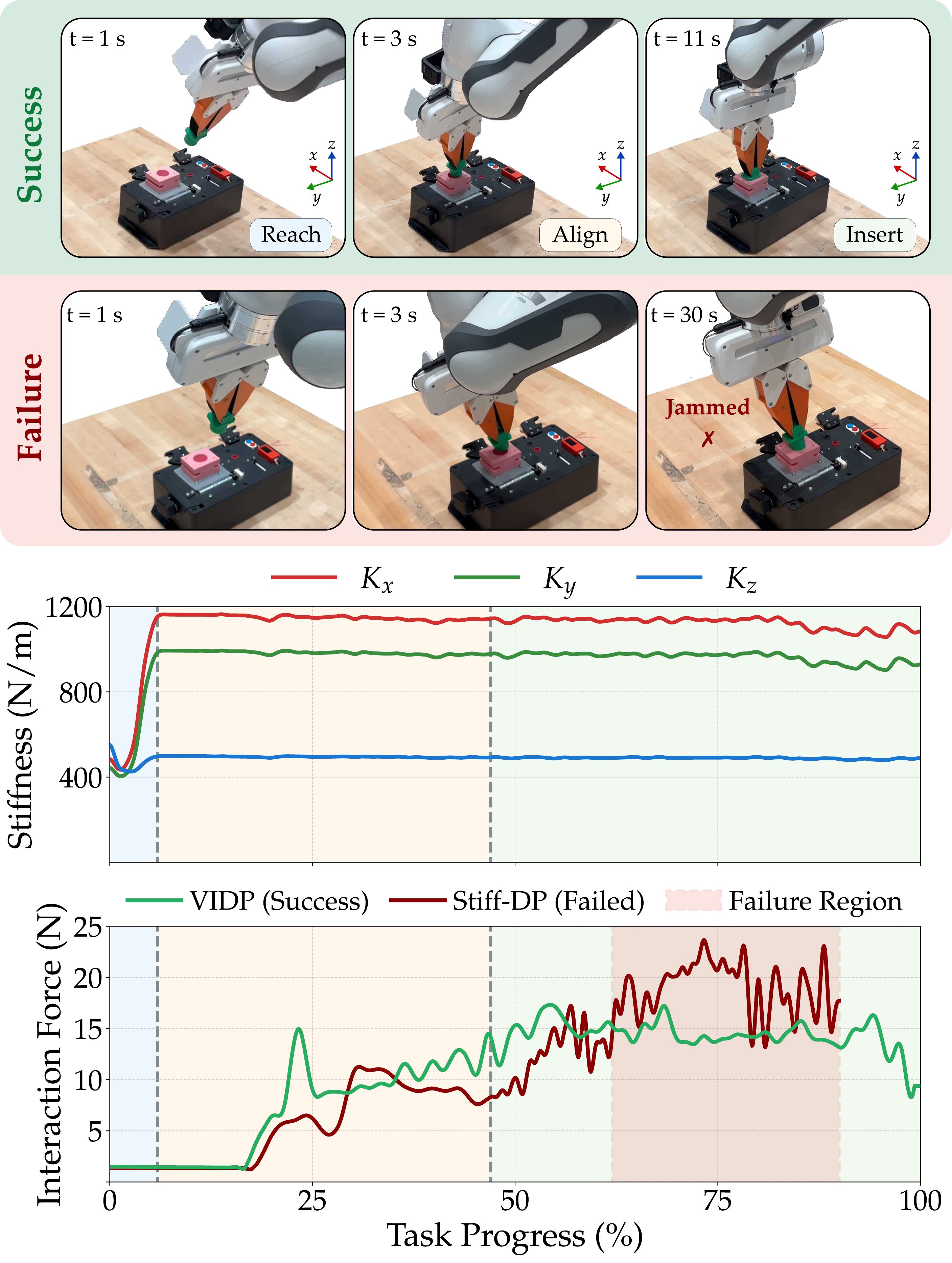}
  \subfiglabel{Peg-in-Hole Insertion}{fig:peg_stiffness}
  \end{minipage}%
  \hspace{0.004\linewidth}%
  \begin{minipage}[t]{0.322\linewidth}
    \centering
    \includegraphics[width=\linewidth]{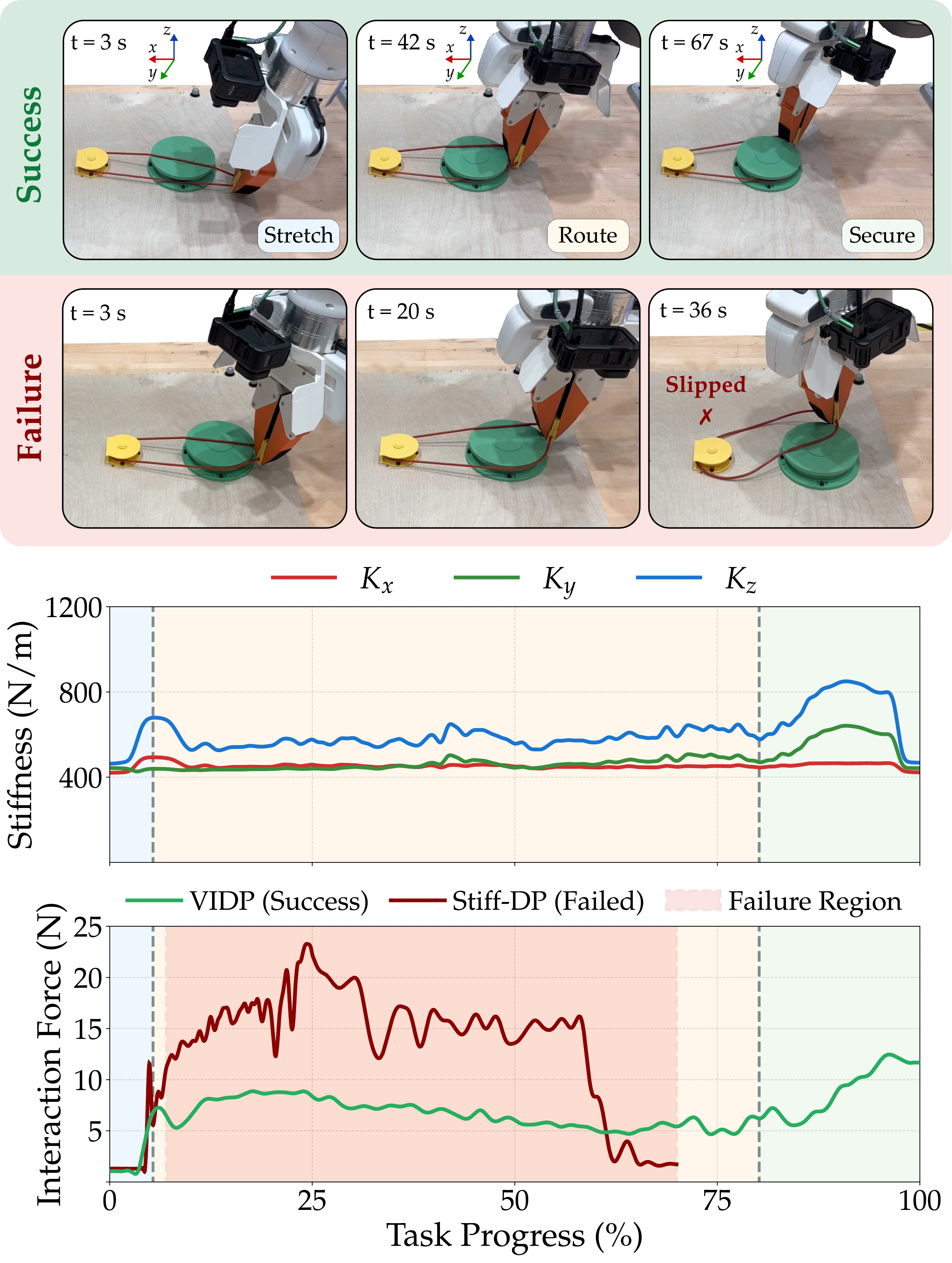}
  \subfiglabel{Pulley Assembly}{fig:pulley_stiffness}
  \end{minipage}%
  \hspace{0.004\linewidth}%
    \begin{minipage}[t]{0.3235\linewidth}
    \centering
    \includegraphics[width=\linewidth]{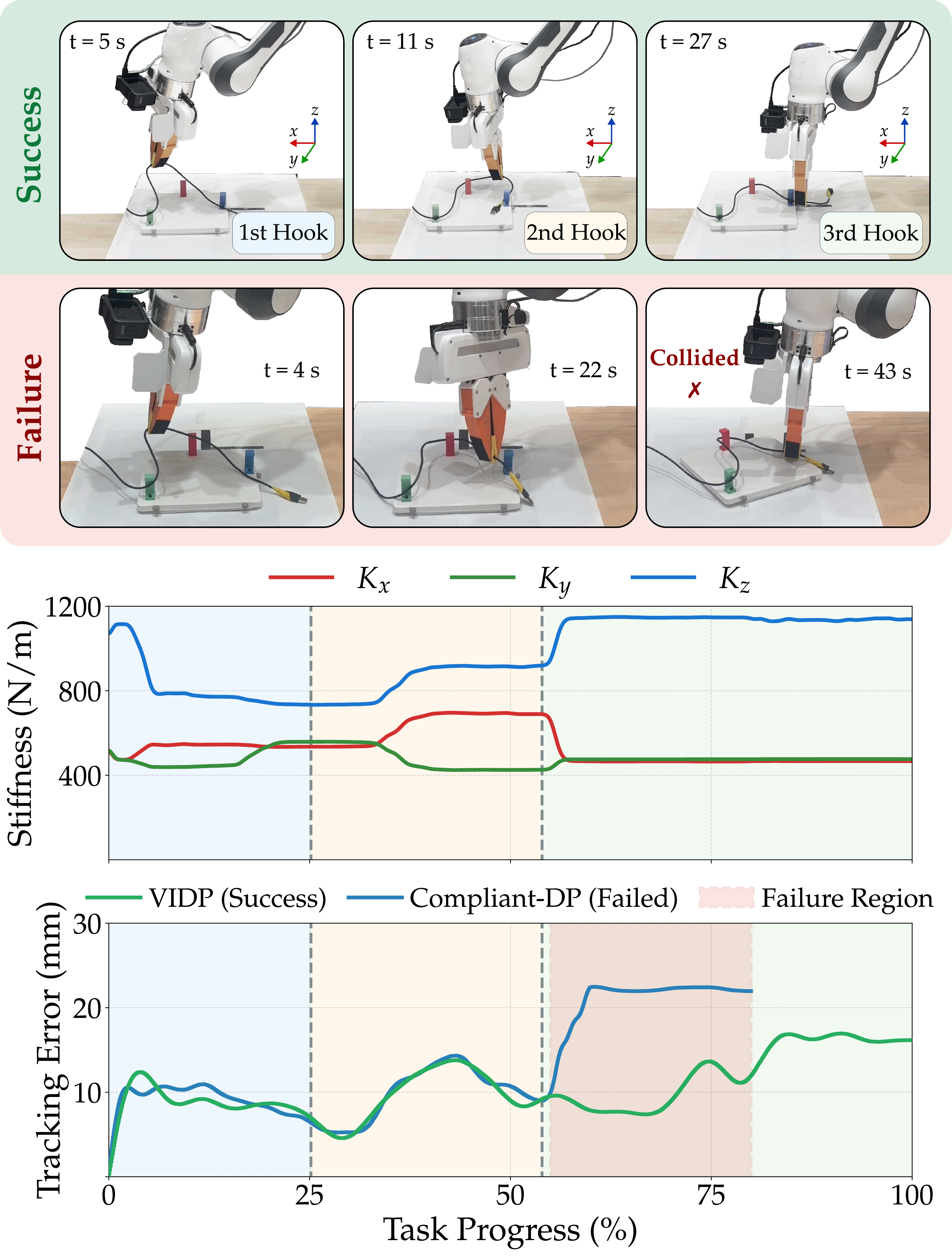}
        \subfiglabel{Cable Routing}{fig:cable_stiffness}
  \end{minipage}
\caption{Phase-wise analysis of one representative successful and failed trial per task. (Top) Snapshots of execution phases completed by VIDP and corresponding failed task phases by Stiff-DP for (a) and (b) and Compliant-DP for (c). (Middle) Cartesian stiffness profiles inferred by VIDP, dynamically modulated to accommodate task constraints. (Bottom) Interaction forces and tracking errors over time. Fixed-impedance baselines experience severe force spikes, causing jamming in (a) \textit{Peg-in-Hole Insertion} and slipping in (b) \textit{Pulley Assembly}, or large tracking errors causing workspace collisions in \textit{Cable Routing}. In contrast, VIDP adapts its compliance to safely and precisely complete all task phases. Experiment trials can be found in the attached video.}
\label{fig:stiffness}
  \vspace{-3mm}
\end{figure*}

\subsubsection{Quantitative Evaluation on Contact-Rich Task Datasets}
To quantitatively evaluate TP-DAMM's clustering directional consistency and spatial preservation on diverse 3D motions, Table \ref{tab:compact_all_task_clustering} summarizes three core metrics across the contact-rich task datasets: directional variance (velocity deviation from the cluster mean, evaluated both in the global frame and locally averaged across task frames), cosine similarity (alignment with the principal direction), and cluster coverage (proportion of demonstrations containing a specific cluster). TP-DAMM consistently achieves the lowest local and global directional variance. This improvement is particularly evident in \textit{Cable Routing} and \textit{Pulley Assembly}, where TP-DAMM significantly minimizes local variance while simultaneously maintaining the highest cosine similarity and full cluster coverage. These results confirm that combining task-frame alignment with direction-aware clustering extracts more physically consistent motion primitives from diverse demonstrations than the baseline methods.

\subsection{VIDP Performance in Contact-Rich Manipulation}
To answer \textbf{RQ2}, we evaluate overall and phase-wise success rates across 20 rollouts per method (Table \ref{tab:success_rate_overall}) with varying spatial workspace layouts. These results motivate a phase-wise analysis of fixed-impedance failures, in which rigid tracking generates excessive forces during constrained contact, while compliant tracking lacks precision in free space.

\subsubsection{Peg-in-Hole Insertion}
VIDP matches Stiff-DP with an 80\% success rate and substantially outperforms Compliant-DP at 15\% (Table \ref{tab:success_rate_overall}), indicating that alignment and insertion require precise pose tracking achieved by high stiffness and degraded by lower stiffness. However, Stiff-DP can still fail through rotational misalignment and jamming due to excessive forces, as shown in Fig. \figsubref{fig:stiffness}{fig:peg_stiffness}. VIDP achieves the same overall success while attaining a slightly higher \textit{Reach} phase success rate (90\% vs. 85\%). As shown in Fig. \figsubref{fig:stiffness}{fig:peg_stiffness}, VIDP maintains high $K_x$ and $K_y$ for accurate tracking while reducing $K_z$ during insertion to regulate contact forces and avoid the jamming observed with Stiff-DP.

\subsubsection{Pulley Assembly}
VIDP achieves 70\% success, nearly doubling Compliant-DP (40\%) and Stiff-DP (35\%). While all methods succeed during the initial \textit{Stretch} phase, Compliant-DP falls behind during \textit{Route} (40\%), and both fixed-impedance baselines struggle to \textit{Secure} the band: Stiff-DP causes slipping, while Compliant-DP suffers from jamming or insufficient fitting authority. As shown in Fig. \figsubref{fig:stiffness}{fig:pulley_stiffness}, Stiff-DP generates excessive forces during constrained routing, destabilizing the band and causing it to slip during the \textit{Secure} phase. In contrast, VIDP applies higher $K_z$ during stretching for vertical alignment, then moderately increases $K_y$ and $K_z$ during \textit{Secure} to fit the band around the pulley.

\subsubsection{Cable Routing}
Unlike the previous tasks, \textit{Cable Routing} primarily requires precise manipulation of a deformable object. VIDP achieves 75\% success, outperforming Compliant-DP (60\%) and Stiff-DP (0\%). Stiff-DP fails entirely at the highly constrained 3rd hook due to excessive pulling forces that displace the base or cause the hook to be missed. Compliant-DP better tolerates cable deformation, achieving 90\% success on the first two hooks, but lacks the precision to secure the 3rd hook caused by large tracking errors (see Fig. \figsubref{fig:stiffness}{fig:cable_stiffness}). Therefore, Compliant-DP led to frequent collisions with the base. On the other hand, VIDP initializes the controller with high vertical stiffness ($K_z$) to fit the 1st hook, then lowers $K_y$ to yield to resistance during lateral motion. It increases $K_x$ for lateral precision in the \textit{2nd Hook} phase and raises $K_z$ in the \textit{3rd Hook} phase to maintain the height constraint. Ultimately, VIDP incurred only one collision failure.

\begin{table}[!h]
\centering
\caption{\textsc{Overall and phase-wise task success rates across 20 rollouts per method ($\uparrow$ is better). Best results are bolded.}}
\vspace{-2mm}
\label{tab:success_rate_overall}
\setlength{\tabcolsep}{1.35pt}
\setlength{\fboxsep}{1.35pt}

\newcommand{\ph}[1]{\makebox[2.7em][c]{#1}}
\newcommand{\phcable}[1]{\makebox[2.25em][c]{#1}}
\newcommand{\phpulley}[1]{\makebox[2.7em][c]{#1}}
\newcommand{\ovr}[1]{\makebox[2.7em][c]{#1}}

\resizebox{\columnwidth}{!}{%
\begin{tabular}{@{}l c|ccc c|ccc c|ccc@{}}
\toprule
\multirow{3}{*}{\textbf{Method}} 
& \multicolumn{4}{c}{\textbf{Peg-in-Hole Insertion}} 
& \multicolumn{4}{c}{\textbf{Pulley Assembly}} 
& \multicolumn{4}{c}{\textbf{Cable Routing}} \\
\cmidrule(lr){2-5} \cmidrule(lr){6-9} \cmidrule(lr){10-13}

& \multicolumn{1}{c}{\multirow{2}{*}{\textbf{\ovr{Ovr. $\uparrow$}}}} 
& \multicolumn{3}{c}{\textbf{Phases $\uparrow$}} 
& \multicolumn{1}{c}{\multirow{2}{*}{\textbf{\ovr{Ovr. $\uparrow$}}}} 
& \multicolumn{3}{c}{\textbf{Phases $\uparrow$}} 
& \multicolumn{1}{c}{\multirow{2}{*}{\textbf{\ovr{Ovr. $\uparrow$}}}} 
& \multicolumn{3}{c}{\textbf{Phases $\uparrow$}} \\
\cmidrule(lr){3-5} \cmidrule(lr){7-9} \cmidrule(lr){11-13}

& \multicolumn{1}{c}{} 
& \ph{Reach} & \ph{Align} & \ph{Insert} 
& \multicolumn{1}{c}{} 
& \phpulley{Stretch} & \phpulley{Route} & \phpulley{Secure}
& \multicolumn{1}{c}{} 
& \phcable{1st} & \phcable{2nd} & \phcable{3rd} \\ 

\midrule
Compliant-DP 
& \ovr{15\%} & \ph{75\%} & \ph{15\%} & \ph{15\%} 
& \ovr{40\%} & \phpulley{90\%} & \phpulley{40\%} & \phpulley{40\%}
& \ovr{60\%} & \phcable{90\%} & \phcable{\textbf{90\%}} & \phcable{60\%} \\

Stiff-DP 
& \ovr{\textbf{80\%}} & \ph{85\%} & \ph{\textbf{85\%}} & \ph{\textbf{80\%}} 
& \ovr{35\%} & \phpulley{\textbf{100\%}} & \phpulley{\textbf{90\%}} & \phpulley{35\%}
& \ovr{0\%} & \phcable{75\%} & \phcable{65\%} & \phcable{0\%} \\ 

\midrule
\rowcolor{gray!12}
VIDP (Ours) 
& \ovr{\colorbox{blue!15}{\textbf{80\%}}} & \ph{\textbf{90\%}} & \ph{80\%} & \ph{\textbf{80\%}} 
& \ovr{\colorbox{blue!15}{\textbf{70\%}}} & \phpulley{\textbf{100\%}} & \phpulley{70\%} & \phpulley{\textbf{70\%}}
& \ovr{\colorbox{blue!15}{\textbf{75\%}}} & \phcable{\textbf{95\%}} & \phcable{\textbf{90\%}} & \phcable{\textbf{75\%}} \\ 

\bottomrule \\
\multicolumn{13}{@{}l}{\footnotesize Ovr. denotes the overall task success rate.} \\
\multicolumn{13}{@{}l}{\footnotesize 1st, 2nd, and 3rd denote the 1st Hook, 2nd Hook, and 3rd Hook phases of the Cable Routing task.}
\end{tabular}%
}
\end{table}

\begin{figure}[t]
  \centering
  \begin{minipage}[t]{0.47\linewidth}
    \centering
    \includegraphics[width=\linewidth]{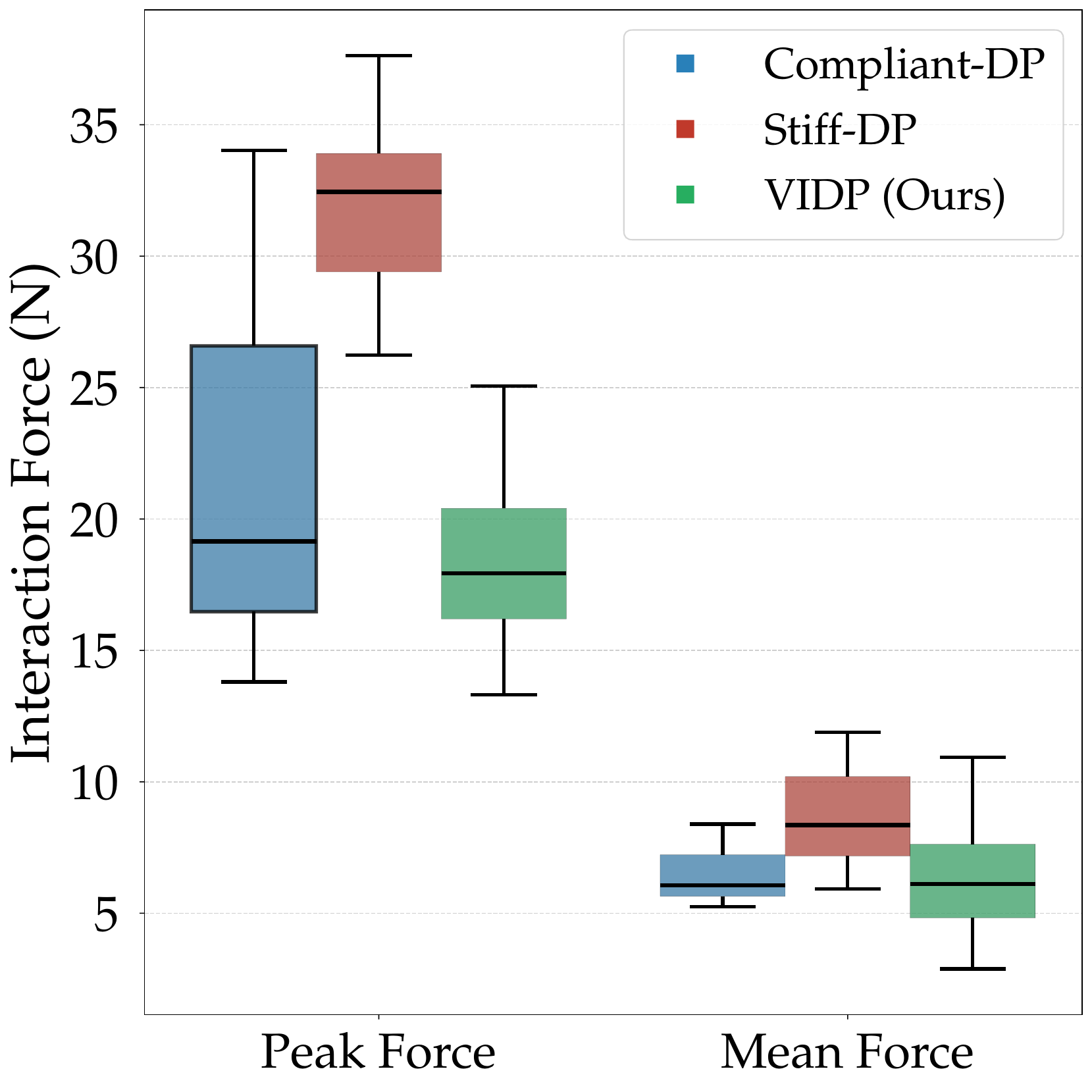}
  \subfiglabel{Peg-in-Hole Insertion}{fig:peg_force}
  \end{minipage}
  \begin{minipage}[t]{0.47\linewidth}
    \centering
    \includegraphics[width=\linewidth]{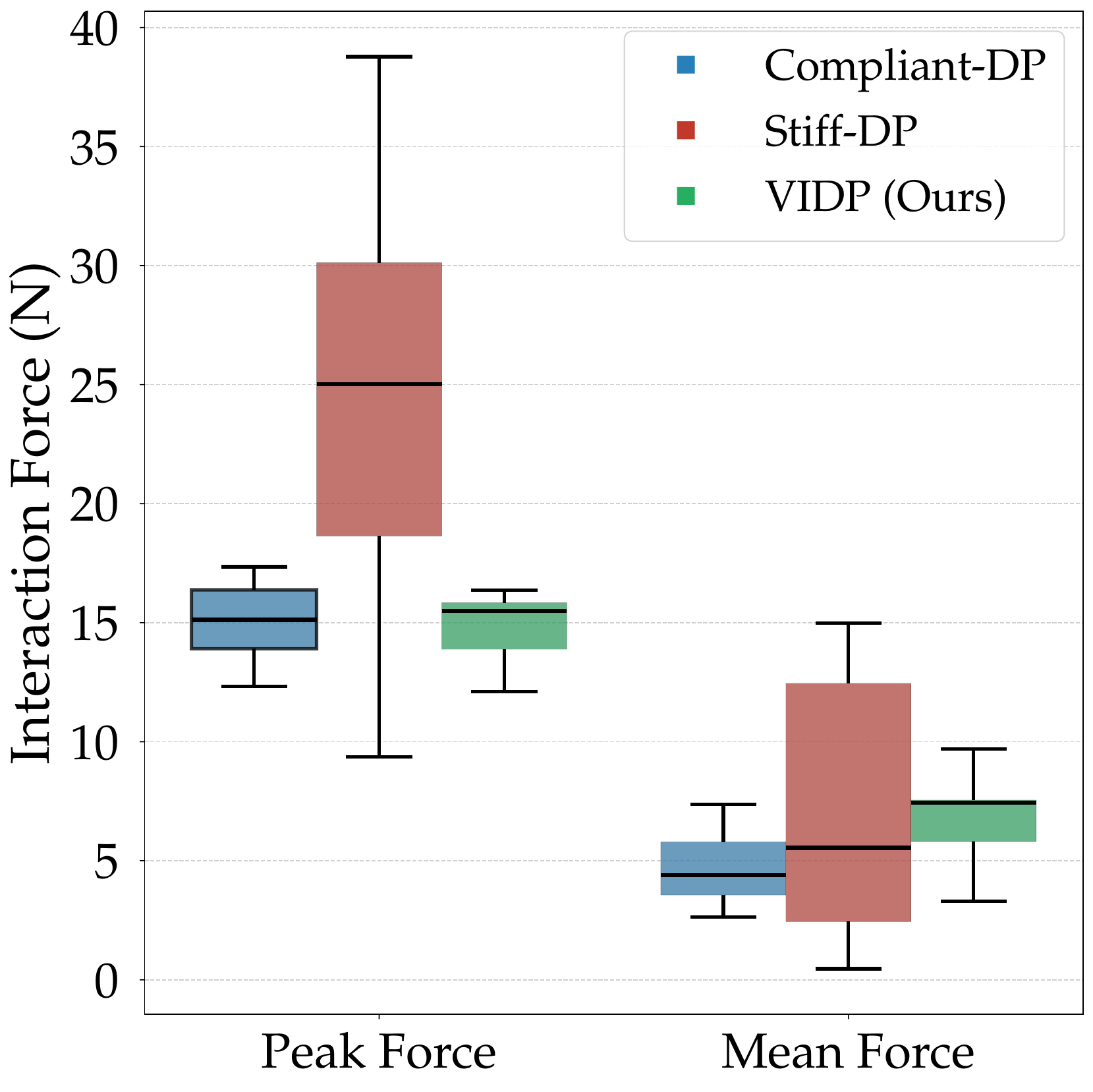}
  \subfiglabel{Pulley Assembly}{fig:pulley_force}
  \end{minipage}
\caption{Peak and mean interaction forces during successful trials of (a) \textit{Peg-in-Hole Insertion} and (b) \textit{Pulley Assembly}. While Stiff-DP provides the necessary tracking precision for these tasks, it generates high and unpredictable force spikes during rigid contact. By modulating stiffness based on the task's physical constraints, VIDP keeps interaction forces within a significantly lower and narrower range, ensuring safer manipulation.}  \label{fig:interaction_forces}
\end{figure}

\begin{table}[t]
\centering
\caption{\textsc{Mean $\pm$ Std Tracking Error (mm) for Successful Cable Routing Trials ($\downarrow$ is better). Best results are bolded.}}
\vspace{-2mm}
\label{tab:tracking_error_cable}
\setlength{\tabcolsep}{4.7pt}
\begin{tabular}{l|cccc}
\toprule
\textbf{Method} 
& \textbf{Total Error} $\downarrow$
& $\boldsymbol{x}$\textbf{-axis} $\downarrow$
& $\boldsymbol{y}$\textbf{-axis} $\downarrow$
& $\boldsymbol{z}$\textbf{-axis} $\downarrow$ \\ 
\midrule
Compliant-DP 
& $10.4 \pm 2.3$ & $4.3 \pm 1.1$ & $8.0 \pm 2.4$ & $2.3 \pm 0.6$ \\
\midrule
\rowcolor{gray!12}
VIDP (Ours) 
& $\mathbf{8.7 \pm 1.7}$ & $\mathbf{3.6 \pm 0.6}$ & $\mathbf{6.9 \pm 1.8}$ & $\mathbf{1.6 \pm 0.2}$ \\ 
\bottomrule 
\end{tabular}
\end{table}

\begin{figure}[t]
  \centering
  \begin{minipage}[t]{0.33\linewidth}
    \centering
    \includegraphics[width=\linewidth]{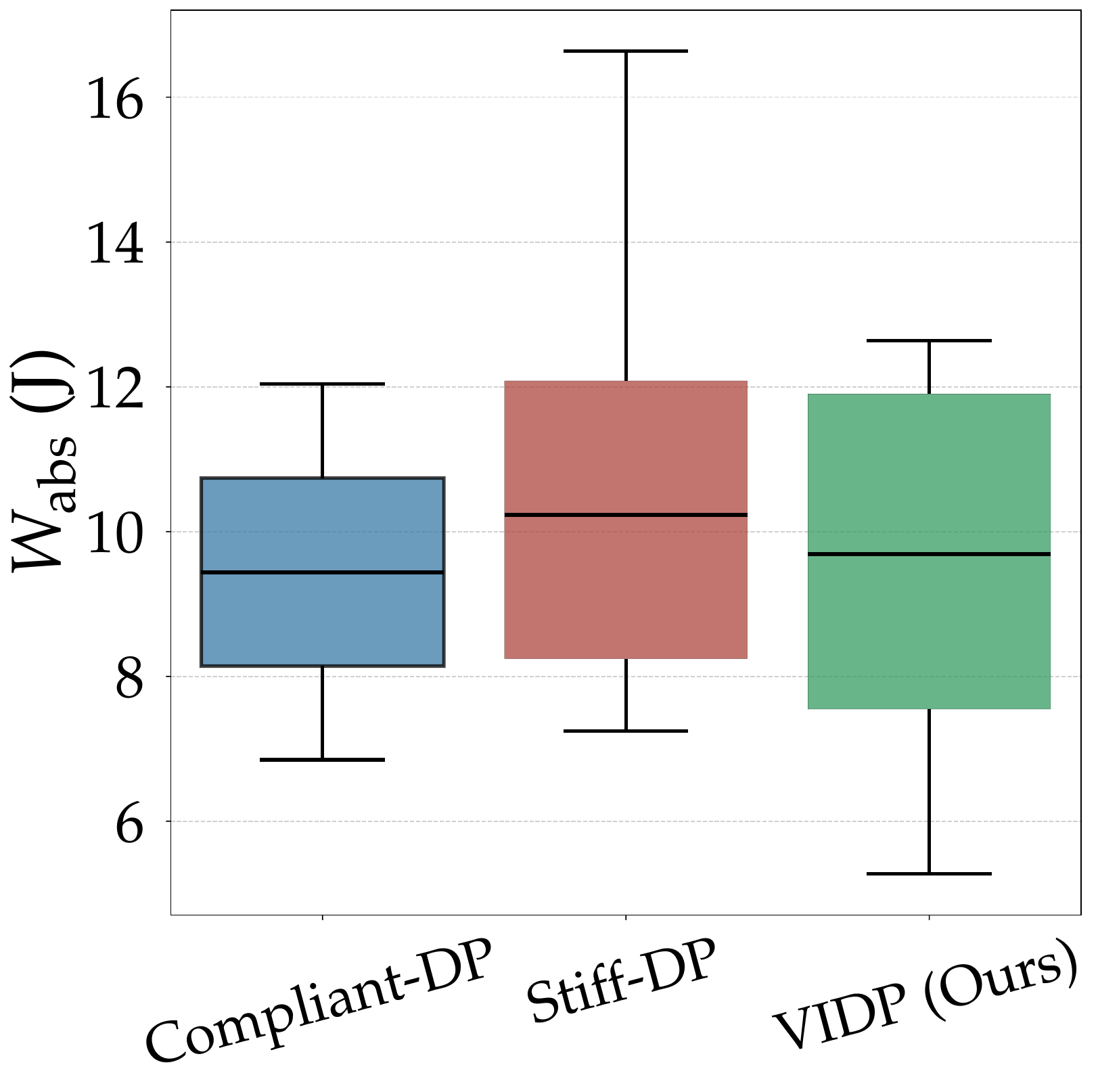}
  \subfiglabel{Peg-in-Hole Insertion}{fig:peg_energy}
  \end{minipage}%
  \hspace{0.004\linewidth}%
  \begin{minipage}[t]{0.33\linewidth}
    \centering
    \includegraphics[width=\linewidth]{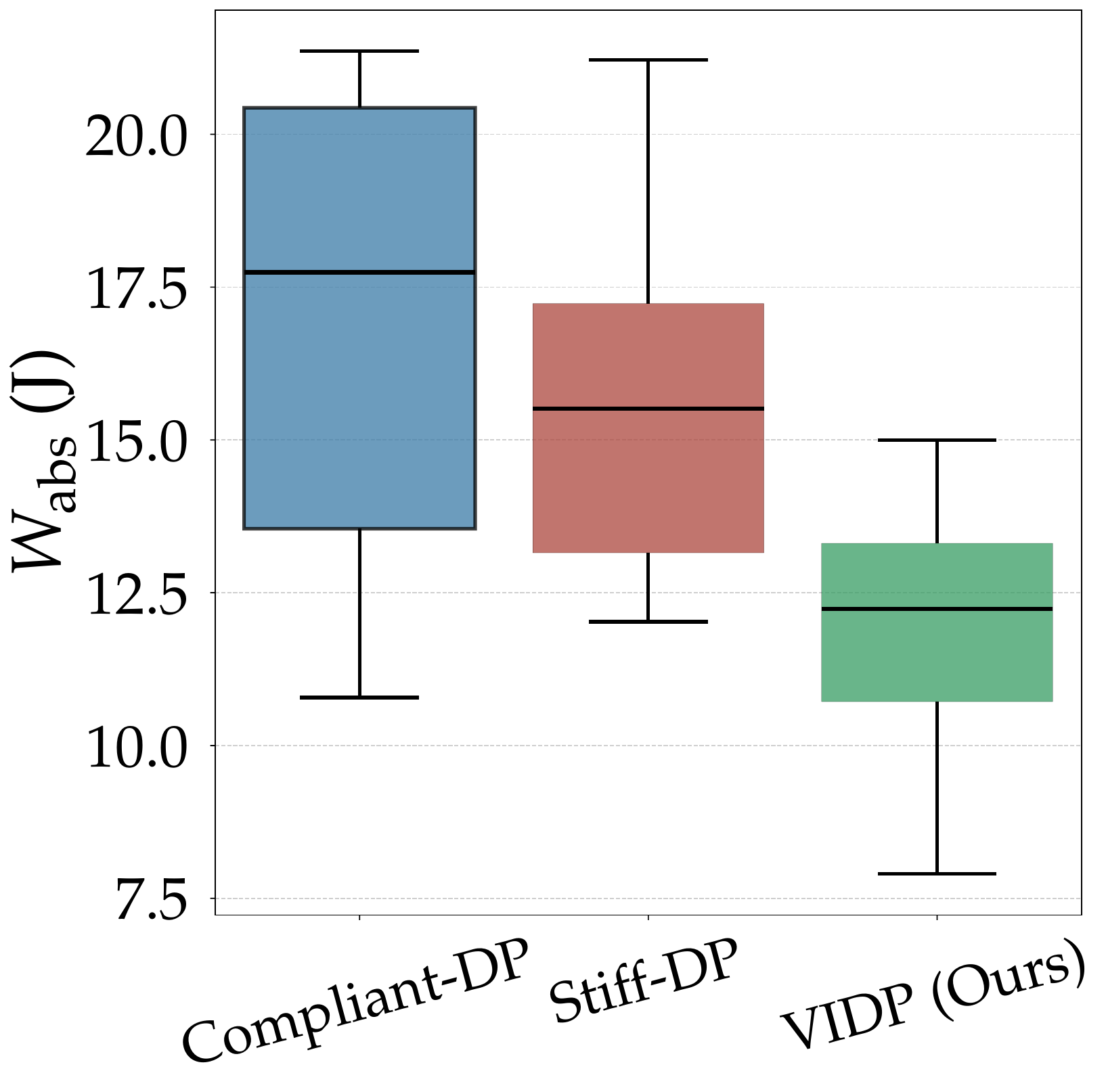}
    \subfiglabel{Pulley Assembly}{fig:pulley_energy}
  \end{minipage}%
  \hspace{0.004\linewidth}%
    \begin{minipage}[t]{0.326\linewidth}
    \centering
    \includegraphics[width=\linewidth]{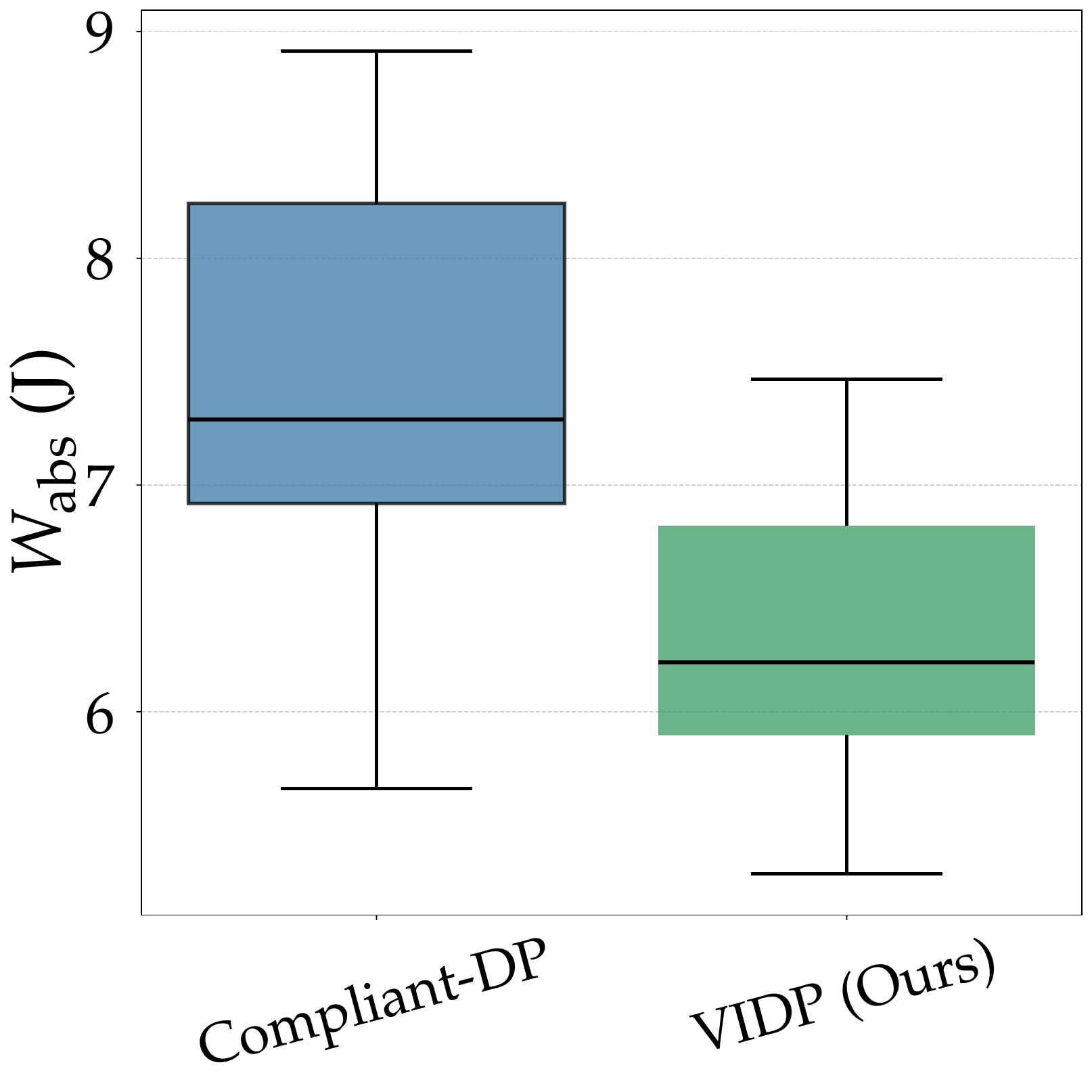}
    \subfiglabel{Cable Routing}{fig:cable_energy}
  \end{minipage}
\caption{Energy efficiency across successful trials measured by absolute mechanical joint work ($W_{abs}$) in Joules. Stiff-DP consumes high energy by persistently resisting rigid contact constraints. Compliant-DP wastes energy through repeated corrective motions to compensate for its tracking errors. VIDP minimizes mechanical work by balancing these physical trade-offs.}
\label{fig:energy}
\end{figure}

\subsection{Interaction Dynamics with VIDP}
To address \textbf{RQ3}, we evaluate physical interaction metrics, specifically interaction forces, tracking errors, and energy efficiency, during successful task executions. By isolating successful trials, we can analyze the inherent physical trade-offs and control behaviors required by each method to complete the tasks.

\subsubsection{Interaction Forces}
Although Stiff-DP provides the rigid tracking needed for rigid contact-based tasks such as \textit{Peg-in-Hole Insertion} and \textit{Pulley Assembly}, it compromises safety and predictability. As shown in Fig. \ref{fig:interaction_forces}, it produces substantially higher peak interaction forces than VIDP for both tasks, even during successful rollouts. In \textit{Pulley Assembly}, the median of the mean force is slightly lower, but its large variance and extreme upper values indicate unpredictable force spikes during contact (Fig. \figsubref{fig:interaction_forces}{fig:pulley_force}). In contrast, VIDP modulates its stiffness to the task's physical constraints, therefore keeping both peak and mean forces within a narrower and more predictable range that better protects the robot and workpiece in contact-rich manipulation.

\subsubsection{Tracking Errors}
In the \textit{Cable Routing} task, maintaining precise tracking is critical even when yielding to cable tension. Table \ref{tab:tracking_error_cable} demonstrates that among successful trials, Compliant-DP results in higher tracking error in all Cartesian directions due to its constant low stiffness. Conversely, VIDP adapts its stiffness over the task phases to maintain tighter precision, ensuring it stays within the reasonable spatial tolerances required for routing the cable around the hooks while retaining enough compliance to smoothly manipulate the cable.

\subsubsection{Energy Efficiency}
VIDP also improves energy efficiency across successful trials, measured by absolute mechanical joint work \cite{thibault2022standardized}. As shown in Fig. \ref{fig:energy}, Stiff-DP requires greater work because it persistently resists rigid contact constraints. In contrast, Compliant-DP consumes the most energy in \textit{Pulley Assembly} and \textit{Cable Routing} (Figs. \figsubref{fig:energy}{fig:pulley_energy} and \figsubref{fig:energy}{fig:cable_energy}, respectively), where low stiffness yields excessively to elastic resistance and requires repeated corrective motions to compensate for tracking errors. VIDP avoids both the high-force effort of Stiff-DP and the inefficient corrections of Compliant-DP, achieving successful executions with consistently lower energy across all tasks.

\section{Conclusion}
\label{sec:conclusion}
This letter proposes Variable Impedance Diffusion Policy (VIDP), a force-agnostic variable impedance control framework based on imitation learning for compliant robot manipulation. By leveraging a novel Task-Parameterized Directionality-Aware Mixture Model (TP-DAMM), VIDP extracts physically consistent trajectory distributions from spatially diverse kinematic demonstrations to predict task stiffness profiles without force sensors. Evaluations across three complex contact-rich tasks validate that VIDP outperforms fixed-impedance baselines in task success rates by balancing safety and precision, ultimately minimizing interaction forces, mechanical work, and tracking errors. Limitations include TP-DAMM's reliance on predefined task reference frames and vulnerabilities to self-intersecting state spaces, alongside VIDP's struggle to handle visually occluded contacts. Therefore, future work will explore methods to estimate variability from more complex task trajectories and to enhance policy reactivity under partial observability.

\bibliographystyle{IEEEtran}
\bibliography{References}
\end{document}